\documentclass[letterpaper, 10 pt, conference]{ieeeconf}  

\IEEEoverridecommandlockouts                              

\let\labelindent\relax
\usepackage{mathptmx} 
\usepackage{times} 
\usepackage{amsmath} 
\usepackage{amssymb}  
\usepackage{graphicx}
\usepackage{subcaption}
\usepackage{caption}
\usepackage{booktabs}
\usepackage{multirow}
\usepackage{float}
\usepackage{subcaption}
\usepackage{xcolor}
\usepackage{amsmath}
\usepackage{amssymb}
\usepackage{enumitem}
\usepackage{upgreek}
\usepackage{xcolor}
\usepackage[
  separate-uncertainty = true,
  multi-part-units = repeat
]{siunitx}
\title{\LARGE \bf
 A Spiking Neural Network Emulating the Structure of the Oculomotor System Requires No Learning to Control a Biomimetic Robotic Head}

\author{Praveenram Balachandar and Konstantinos P. Michmizos, Member \textit{IEEE}
\thanks{PB and KM are with the Computational Brain Lab, Department of Computer Science, Rutgers University, New Jersey, USA
        {\tt\small konstantinos.michmizos@cs.rutgers.edu}}%
}

\begin{document}

\maketitle
\thispagestyle{empty}
\pagestyle{empty}

\begin{abstract}
 Robotic vision introduces requirements for real-time processing of fast-varying, noisy information in a continuously changing environment. In a real-world environment, convenient assumptions, such as static camera systems and deep learning algorithms devouring high volumes of ideally slightly-varying data are hard to survive. Leveraging on recent studies on the neural connectome associated with eye movements, we designed a neuromorphic oculomotor controller and placed it at the heart of our in-house biomimetic robotic head prototype. The controller is unique in the sense that (1) all data are encoded and processed by a spiking neural network (SNN), and (2) by mimicking the associated brain areas' topology, the SNN is biologically interpretable and requires no training to operate. Here, we report the robot's target tracking ability, demonstrate that its eye kinematics are similar to those reported in human eye studies and show that a biologically-constrained learning, although not required for the SNN's function, can be used to further refine its performance. This work aligns with our ongoing effort to develop energy-efficient neuromorphic SNNs and harness their emerging intelligence to control biomimetic robots with versatility and robustness.
\end{abstract}

\section{INTRODUCTION}
 
\par 
Covering all ranges of robotics, from structure \cite{RN245} and mechanics \cite{RN252} to perception \cite{RN257}, actuation \cite{RN262, RN263} and autonomy \cite{RN265}, biomimetic robots imitate the nature's design \cite{RN245} and movement \cite{RN263} principles, to perform desired tasks in unstructured environments \cite{RN257}. An orthogonal direction towards biomimesis is to imitate the most advanced biological controller: the brain. This particular type of brain-mimesis often entails the use of neural networks \cite{RN267}, of varying degrees of complexity \cite{RN271} and biological plausibility \cite{RN273, RN272}, that promise advances to Robotics \cite{tang2020reinforcement, RN275} and insights to Brain Science \cite{RN276, RN277}. 

With robots being arguably the sweet spot for neuromorphic artificial intelligence, the emergence of neuromorphic chips  \cite{davies2018loihi,furber2014spinnaker} has spurred interest for a bottom-up rethinking of biomimetic controllers in the form of spiking neural networks (SNN) that seamlessly integrate to non-Von Neumann architectures. We and others have recently proposed brain-inspired SNNs that embed into such chips and solve robotic problems with unparalleled energy-efficiency  \cite{Guangzhi} while promising a robust yet versatile alternative to the brittle inference-based machine learning solutions \cite{tang2020reinforcement, astroLoihi}. The main criticism to neuromorphic computing is that, in the absence of a strong learning algorithm, the current state of the art does not share the same scaling abilities with the mainstream deep learning methods.

Alongside efforts to implement backpropagation algorithms in SNN \cite{RN255, RN28}, most neuromorphic algorithms are indeed simple enough to be trained via variations of spike-timing-dependent plasticity (STDP), a Hebbian-type local learning rule \cite{RN35, RN27}. An alternative direction, that we have started to explore, entails the SNNs to be dictated by the underlying neural connectome associated with the targeted function \cite{Guangzhi, tang2018}. In this paper, we extend this direction by presenting a neuro-mimetic SNN that draws from the connectome of the human oculomotor system and enables our in-house robotic head to track a laser target. We demonstrate how the robotic prototype achieved real-time tracking of the visual target, by coordinating saccadic and pursuit eye movements with neck movements. Given that the structure of any network, be it biological or artificial, serves its function, the SNN's behavior emerged out of the network's topology, without any training.

\section{Methods}

\subsection{The Robotic Head Prototype}
The biomimetic robotic head prototype comprised of two cameras (eyes) and a neck (Fig. 1a). Each camera was mounted on a pan-tilt mechanism, which allowed horizontal and vertical movement of the eyes. The eyes were fixed to a base plate mounted on top of another pan-tilt system, to replicate neck movements. Overall, the robot had 6 degrees of freedom (DOF). The three pan-tilt systems were controlled by Dynamixel AX-12A digital servos driven by an Arduino-based Arbotix-M controller.  The controller relayed servo position deltas to the Arbotix-M robocontroller through a USB serial interface. The deltas were computed based on the location of the laser target in a foveated field of view. The range-of-motion (ROM) for the servos controlling the eyes was restricted to 100 (70) degrees horizontally (vertically) from the center, to keep the eye kinematics within the biologically plausible range. Similar restrictions for biologically plausible ROM were imposed on neck movement.

The target was projected on a wall at 55 cm away from the eyes, using a laser diode mounted on a separate Arduino-controlled pan-tilt system.  The movement profiles of the laser and the eyes were calibrated with respect to a fixed reference position for estimating the tracking accuracy of the proposed controller, by using linear regression analysis libraries to determine a higher order polynomial relation between the target angle and servo position. The laser and servo positions were recorded and mapped to the eye and neck kinematics, as well as the target position.

\begin{figure*}[h]
    \includegraphics[width=\textwidth]{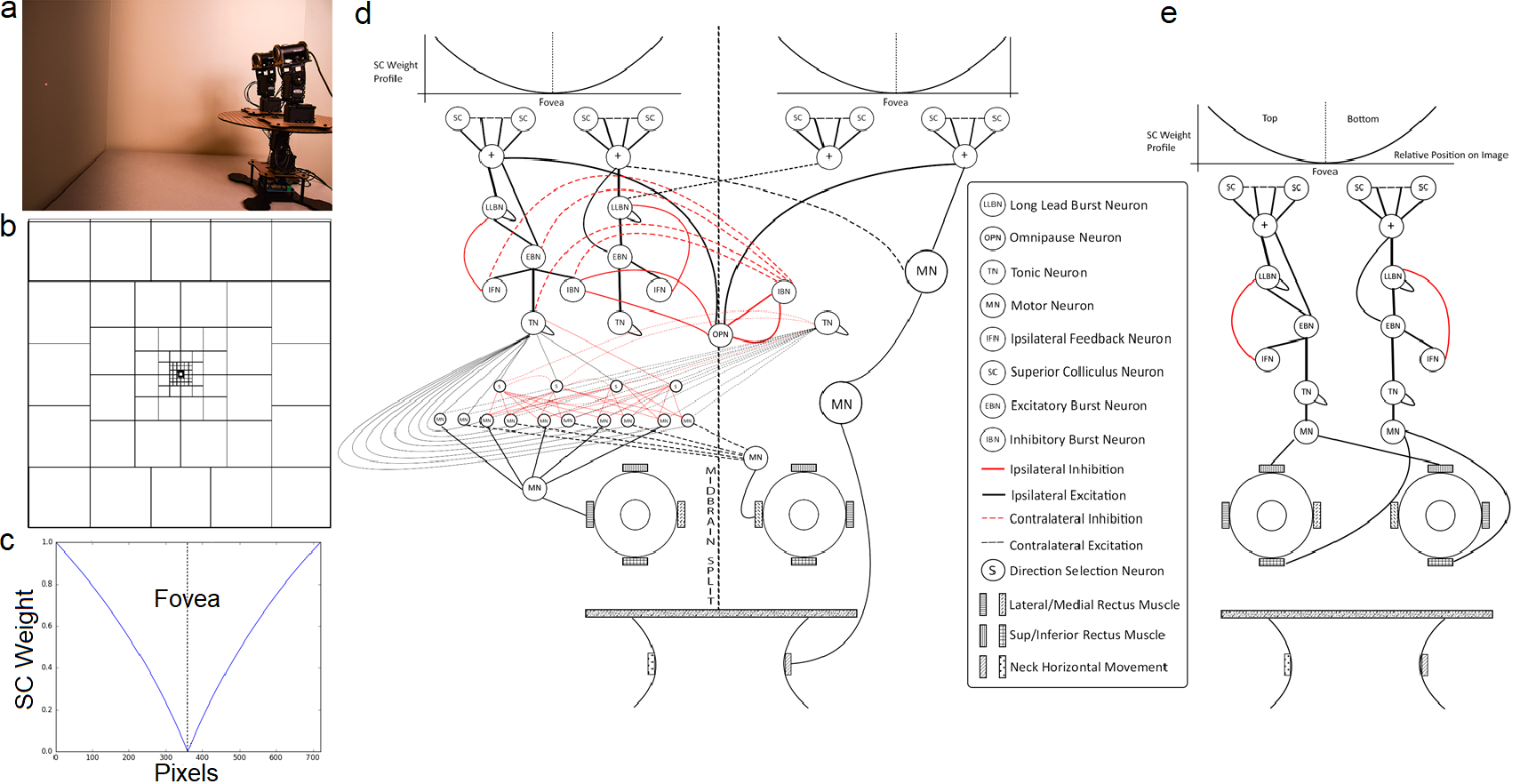}
    \caption{\textbf{(a)} The robotic head prototype fixating at a moving laser target projected on a wall. A pan-tilt system controlled the position of the laser (not shown). \textbf{(b)} The Receptive Field (RF) of the superior colliculus (SC) neurons (artificial retina) shown as squares in the input image. An SC neuron was activated when the input (laser dot) was within their receptive field. \textbf{(c)} The variation of SC neurons' weights with respect to their retinotopic pixel position. The same weight distribution applied to both horizontal and vertical distances. \textbf{(d)} The SNN connectivity for moving the eyes and the neck in the horizontal direction. The shown connectivity represents innervation of the eyes' muscles for moving the eyes to the left and the neck to the right. The opposite movements were controlled by a sub-network of symmetric connectivity to the one shown here (omitted for illustrative purposes). \textbf{(e)} The SNN connectivity for moving the eyes and the neck in the vertical direction. The input SC neurons correspond to the neurons in the top half above the fovea and the bottom half above the fovea for both eyes. The input from the two eyes are used together to generate coordinated movement of both eyes upward or downward.}
    \label{fig:architecture}
\end{figure*}

\subsection{Network Topology for The Oculomotor Controller}

The controller's goal was to keep the moving laser target within the ``fovea'' of each eye through coordinated movements of both eyes and the neck. This was achieved by emulating the associated neurons and brain topology associated with eye and neck movements. In that sense, the SNN's tracking behavior emerged from the network topology and, thereby, required no learning to operate.

\subsubsection{Visual input processing -- Retina and Superior Colliculus}

The robotic head had a foveal vision (Fig. 1b). Each frame drove the input neural layer. Mimicking the retina, the input neuron density decreased with the distance from the center of the frame and the camera's visual field was divided into non-overlapping square receptive fields (RF). Such neurons have been found in the superior colliculus (SC) \cite{prablanc2003single}. The SC neurons exhibited a simple threshold behavior, by firing if there was a target in their RF. Depending on the lighting conditions, an SC neuron could fire when a certain percentage of the pixels were activated within its RF.

The SC neurons innervated the distinct gaze centers and were crucial for the robot's behavior. Their firing rate encoded the amplitude of the eye movements. Specifically, activating peripheral SC neurons resulted to eye movements of larger amplitude compared to when SC neurons with RF close to the fovea were activated. To design this feature, we followed experimental findings \cite{gandhi2011motor} and modeled the weights of the synaptic connections between the SC neurons and the brainstem regions as having an increasing function of the distance (number of pixels) from the fovea (Fig. 1c).

\subsubsection{Oculomotor response generation -- Midbrain and PONS regions}

The SNN topology followed the connectome of the oculomotor system. Specifically, saccade control structures are known to be grouped into two regions for vertical and horizontal movement control, namely  the horizontal gaze center (PPRF) and vertical gaze center (riMLF), found in the pons region of the brainstem \cite{sparks2002brainstem, scudder2002brainstem, scudder1988new}. Likewise, we designed the oculomotor SNN having two separate sub-networks, one controlling horizontal eye and neck movemements (PPRF) and another for the vertical  movements (riMLF) \cite{king1981vertical} (Fig. 1d, e). The SC neurons innervated the two distinct gaze generation centers that controlled vertical and horizontal eye movements separately. The sub-SNN that controlled horizontal movements supported both conjugate and vergence eye movements while the sub-SNN for the vertical control was simpler by moving both eyes simultaneously. 

For vertical eye movements, the SC neurons with RF at the upper (lower) half of the field of view drove the sub-SNNs that generated the up (down) movements. The input from SC neurons was fed to a long lead burst neuron (LLBN) and an excitatory burst neuron (EBN). The LLBN had a delayed excitatory connection with the EBN. The EBN connected to 1) an ipsilateral feedback neuron (IFN), which inhbited the LLBN and thereby controlled the overal response of the SNN, as well as 2) a tonic neuron (TN), with a sustained firing activity that triggered the motor neurons (MN), which drove the servos up (down). As both eyes were controlled by the same network, the robotic eyes moved up (down) by the same amount, to bring the fovea back onto the target.

For horizontal conjugate eye movements, the SC neurons with RF at the left (right) half of the field of view drove the sub-SNN that controlled the left (right) eye movements. The conjugate eye movement is facilitated in the brain through the abducens nuclei and interneurons that trigger the response of the contralateral oculomotor nuclei \cite{sparks2002brainstem, scudder2002brainstem}. Similarly, the input from SC neurons drove the LLBN and the EBN which drove the IFN and TN, as described in the vertical direction, but this time an inhibitory burst neuron (IBN) inhibited the contralateral eye movement through the omnipause neuron (OPN). The OPNs' role was to control fixation and maintain the fixation on a target, by inhibiting the activity of EBNs and IBNs. The IBN also inhibited the activity of the contralateral EBN, IBN, IFN and TN, which ensured that the contralateral eye would not move to the opposite direction (e.g., the right eye would not move to the right when the left eye would be moving to the left). The OPN received inhibitory inputs from both IBNs on both sides, and had an inhibitory feedback connection to the same IBNs. These connections controlled the inhibitory output of the IBN as the eyes moved in the same direction.  The combined effect of excitation from the SC neurons and the selective inhibition from the OPNs and IBNs led to the movement of both eyes in the same direction towards the target. In addition to the ipsilateral TN, additional direction selection neurons (DSN) drove the MNs. These DSNs got excitatory input from the ipsilateral TN, and inhibitory input from the contralateral TN, and drove the network of MNs that controlled the servos. These connections enabled the simultaneous movement of both eyes to one direction, representing, e.g., the lateral and the medial rectus muscle of the left and right eye, respectively. 

For horizontal dis-conjugate (vergence) movements, the sub-SNNs were similarly driven by SC neurons with RF at either the left or the right half of the field of view. Several bursting neurons are found to be direction selective \cite{ohtsuka1990direction}, contributing to the oculomotor response only in a specific direction. We modeled these neurons, denoted by S in Fig. 1, that received inputs from both the ipsilateral and contralateral sides and inhibited the activity that promoted conjugate eye movements. In this case, the EBN ony connected to an IFN and a TN that had the same connectome to the conjugate SNN controller. TN excited the DSN that controlled lateral eye movements, alongside input from the contralateral eye. This mechanism for the independent control of the robotic eyes is also in alignment with experimental findings \cite{scudder2002brainstem}.

For neck movement, we followed experimental findings and designed a simple sub-SNN so that only when the target was significantly away from the fovea, the controller engaged the neck \cite{corneil2002neck}. To do so, we designed the SC neurons from the periphery of the visual field to drive directly the MNs that controlled the neck in the four directions. The SC-MN connection strengths were such that the neck would not move when only the eye movements were sufficient to bring the target inside the fovea.

\subsubsection{Motor response output generation}
The effective output of the sub-networks for horizontal and vertical movement was translated into equivalent servo position deltas using a firing rate encoding scheme. The firing rate was computed over windows of ~20ms and then scaled to a servo position value between 0 to 1023, which was the range of valid positions for the AX-12A servos. 

\subsection{Reward-based Hebbian Learning}

Although the SNN can control the robotic head with no training, we included a biologically plausible learning mechanism to examine whether training could refine the robot's performance. Interestingly, saccadic amplitude in primates is known to adapt to both a bottom-up visual error signal \cite{wallman1998saccadic} and a top-down behavioral (goal) signal \cite{schutz2014saccadic}. Here, we introduced a bottom-up reward-based learning mechanism, based on Sejnowski’s Hebbian learning rule \cite{sejnowski1977storing}.

Reward-based learning relies on maximizing the reward signal associated with the network performing as expected, here when the target was on the fovea. A global reward signal promoted the SNN behavior in bringing the target onto the fovea. The synaptic adaptation was semi-local, i.e. the weight change depended on the global reward signal and the pre-synaptic and post-synaptic neuron activities at the synapse. The reward value as a function of the position of the target on the frame is shown in Fig. 2. For every pair of pre-synaptic neuron $j$ and post-synaptic neuron $i$, reward-based Hebbian learning was defined as: 

\begin{figure}[t]
    \centering
    \includegraphics[width=\linewidth]{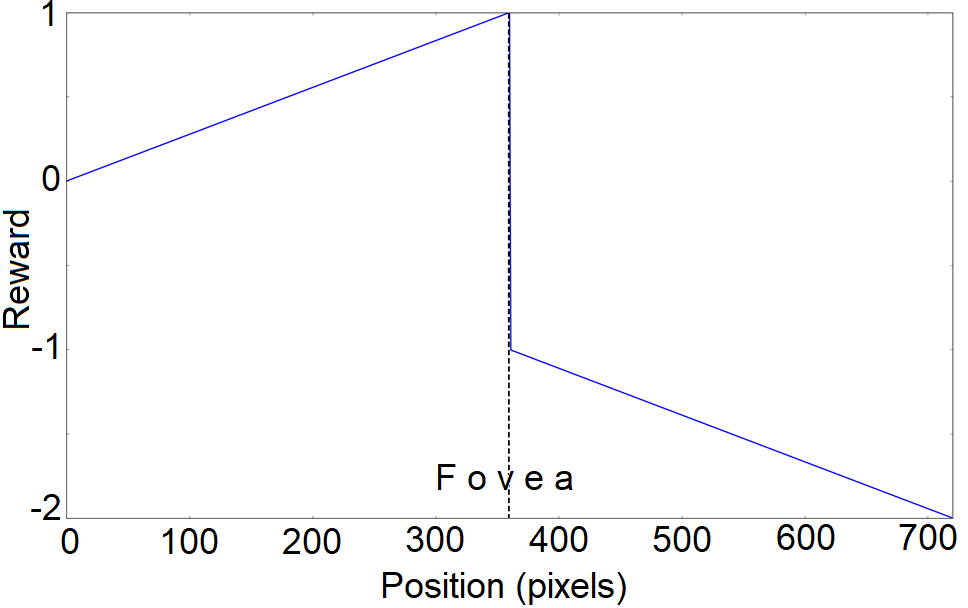}
    \caption{Reward value as a function of the target position. The fovea center is at position = 360 pixels.}
    \label{fig:reward}
\end{figure}

\begin{equation}
    {\uptau}_e \frac{de_{ij}}{dt} = - e_{ij} + H(pre_j, post_i)
\end{equation}
\begin{equation}    
    \frac{dw_{ij}}{dt} = M H(pre_j, post_i) e_{ij}
\end{equation}    
\begin{equation}
    M(t) = R(t)- <R>
\end{equation}

where $e_{ij}$ is the synaptic eligibility trace for the pair of neurons, $w_{ij}$ is the weight of the synapse between those neurons, $H$ is the Hebbian learning term and $M(t)$ is the neuromodulator signal at time $t$ denoting the difference between given and expected rewards. Here, we empirically estimated the expected reward $<R>$ as the running average and the time constant ${\tau}_e$ was chosen in the range of 1 sec, to bridge the delay between action choice and final reward signal. The Hebbian term $(H(pre_j, post_i))$, was modelled based on the Sejnowski learning rule. This rule relies on the activities of pre- and post-synaptic neurons, as well as the rate of spikes of these neurons over a window. It is based on the idea that firing rate of the neurons vary around their mean values $<v_i>$ and $<v_j>$ and defined by:

\begin{equation}    
    \frac{dw_{ij}}{dt} = \gamma (v_i - <v_i>) (v_j - <v_j >)
\end{equation}    

where $\gamma$ is the learning rate and $v_{j,i}$ are the firing rates of the pre- and post-synaptic neuron, respectively.

The laser and the servo positions for each DOF of the robotic head was recorded at 45 Hz and the kinematics of the eyes with respect to that of the laser was studied with and without learning. Both repetitive and random pattern of target positions were used to train the controller and the mean accuracy was used as a measure of behavioral performance. 

\section{Results}

We validated the proposed SNN for its ability to track accurately a moving target, both without and with Hebbian learning. When we did not incorporate learning into the SNN, the vast majority of its weights were adapted by the relative strengths of the connections between neural areas found in experimental studies, and the rest weights were found by trial and error. The C++ code is available in Appendix B in \cite{praveen}. For the kinematics, the sign of the angles represented position of the target with respect to the origin, defined as the center of the frame. For horizontal movement, the negative (positive) angles represented positions on the left (right) side of the origin. Similarly, for vertical movement, negative (positive) values represented positions below (above) the origin. For calculating the discrepancy between fovea and target positions, the neck position was added to the horizontal and vertical components of each eye.

\begin{figure}[t!]
    \centering
    \includegraphics[width=\linewidth]{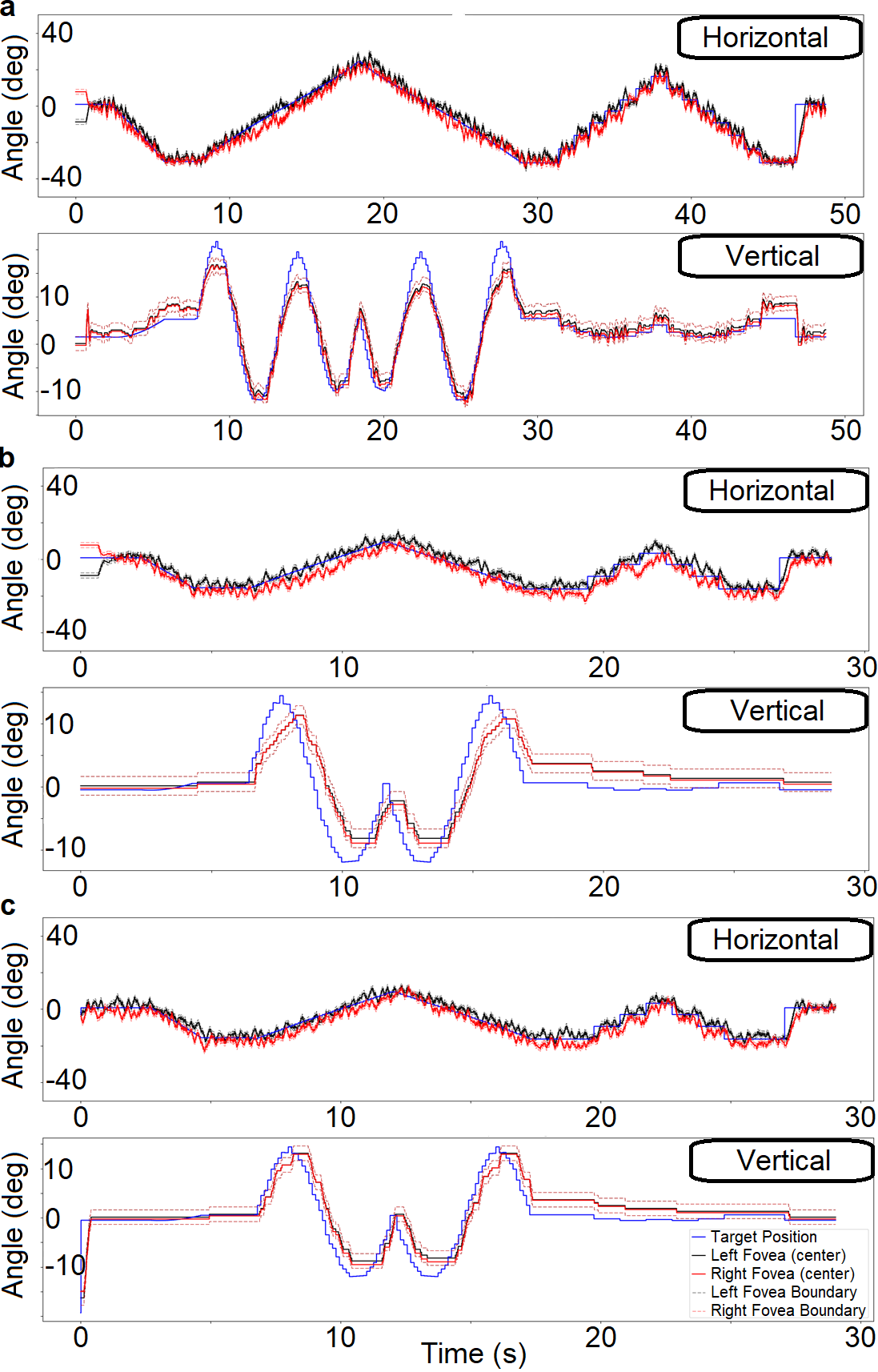}
    \caption{(a) Eye kinematics with respect to target kinematics. The target, shown in blue, is moved around through a sequence of positions on the wall demonstrating sudden changes in both horizontal and vertical directions. The eyes, shown in red and black, follow the target by making multiple saccade like movements. Top panel shows the horizontal kinematics of the eyes with respect to the target and bottom panel shows the vertical component of the movement.  (b)  Kinematics of the eyes for a sequence of target positions without learning. (c) Kinematics of the eyes for a sequence of target positions with reward-based hebbian learning.}
    \label{fig:behavior}
\end{figure}

For the no-learning SNN, the mean relative error for the  eye position with respect to the target, averaged over both horizontal and vertical direction of both the eyes and neck, was $-0.685^{\circ}$, averaged over ten 2-minute tracking experiments, for randomly moving targets. An example of the kinematics of the prototype with respect to the kinematics of the laser target on the wall is shown in Fig. 3a. We observed small, jerk-like, eye movements, similar to the experimentally reported miniature versions of voluntary saccades, named microsaccades. We attribute this behavior to the property of the SNN controller, which tries to fixate on the target within the bounds of the fovea: When the dimensions of the fovea were smaller than the target size, the controller constantly readjusted itself, attempting to fit the target within the fovea. This oscilatory activity was particularly evident in the horizontal position, partly because of the larger complexity of the underlying network, compared to the one controlling the vertical movements. Interestingly, the vertical movement exhibited a delayed response (lower bandwidth), leading to lower accuracy for vertical tracking compared to horizontal tracking. This was further corrected by allowing reinforcement learning.

The introduction of reward-based learning into the SNN resulted to a better tracking ability (Fig. 3c). Learning allowed for a noticeable improvement in tracking for targets with fast vertical components. To quantify the tracking abilities in both cases, we presented a random sequence of target positions to both controllers (Table I.) 

\begin{table}[t!]
      \centering
  \begin{center}
    \caption{Median Relative Error (RE) and root mean square  error (RMSE) for eye position with respect to target, without and with Hebbian learning (-HL), for ten 2-min experiments where the robot tracked randomly moving targets.}
    \label{tab:lot}
    \begin{tabular}{l|r|r|r|l}
      \toprule 
      \textbf{Eye Kinematics} & \textbf{RE} & \textbf{RE-HL} & \textbf{RMSE} & \textbf{RMSE-HL}\\
      \midrule 
      Left Eye - Horizontal & $-1.87^{\circ}$ & $-1.487^{\circ}$ & $3.55^{\circ}$ & $2.87^{\circ}$ \\
      Right Eye - Horizontal & $2.115^{\circ}$ & $2.049^{\circ}$ & $3.93^{\circ}$ & $3.44^{\circ}$ \\
      Left Eye - Vertical & $-1.219^{\circ}$ & $-0.697^{\circ}$ & $3.14^{\circ}$ & $3.02^{\circ}$ \\
      Right Eye - Vertical & $-0.823^{\circ}$ & $-0.421^{\circ}$ & $3.03^{\circ}$ & $2.95^{\circ}$ \\
      \bottomrule 
    \end{tabular}
  \end{center}
\end{table}

\section{Discussion}

Here, we introduced a biologically interpretable SNN oculomotor controller and its integration to our in-house robotic head. The kinematics of the robot's eyes resembled those of the human gaze in tracking a moving target \cite{schutz2014saccadic}. In that sense, the goal of this work was to demonstrate that ``machine behavior'' in general, and robotic function in particular, can emerge naturally from an \textit{a-priori} knowledge that dictates or informs network topology. By drawing from the connectome of the brain areas associated with the targeted behavior, this and other efforts to develop biologically realistic SNNs can help advance Brain Science and Robotics, two fast growing fields that are also converging via biomimicry and neuromorphic computing \cite{RN217, RN54, Guangzhi, tang2018}.

In the technical domain, adding biological constraints to an SNN structure removes the need for assuming all-to-all initial connectivity for the trainable network. This may translate to further improvements in training efficiency, as it limits learning to a small number of synaptic connections. In addition, contrary to the typical neuron models in deep networks that can be optimized to perform complex computational tasks \cite{RN4}, spiking neuron models have a non-differentiable output (their all-or-none firing) and therefore are incompatible with standard gradient-descent supervised learning methods \cite{RN201}. In the absence of a strong learning algorithm, the main criticism to neuromorphic solutions is that promising preliminary results \cite{Guangzhi} cannot share the same scaling abilities with the mainstream deep learning approaches. Here we show the first fruits of our efforts towards scalable SNNs that, by being able to host biological principles of computation known to be critical for intelligence, can give end-to-end neuromorphic solutions towards fully autonomous systems.

Effective as they may have become, robots still cannot duplicate a range of human behaviors, such as dynamically responding to changing environments using error-prone sensors. To operate in a real-world environment, an autonomous robot should 1) be robust to a noisy neural representation, 2) adapt to a fast changing environment, and 3) learn with no or limited supervision or reinforcement. The embodiment of SNNs into robots has been rather sparse and the current approaches aim to give a proof of concept \cite{RN270, RN278, RN274}, rather than a whole-behaving robot. While there is definitely value in studying simplified tasks and basic sensory representations \cite{RN279}, there is an ongoing need to propose new controllers capable of naturally handling richer, noisier and more complex scenarios \cite{RN218}. This work suggests that a promising path towards duplicating a human-like behavior is to draw knowledge from how synergy is achieved in neurons across the implicated brain areas. In addition, contrary to the constantly online processing taking place in the oculomotor system, the traditional learning algorithms rely on separate training and inference phases. That is why we employed unsupervised learning, which is a better fit for lifelong learning of a continuously evolving network that can adapt to new targets and movement patterns. 

The re-emergence of neuromorphic computing calls for a bottom-up rethinking of computational algorithms that can seamlessly integrate into non-Von Neumann hardware \cite{Guangzhi}, promising unparalleled energy-efficiency and a robust yet versatile alternative to the brittle inference-based AI solutions \cite{RN298}. This work brings us closer to realize this promise by tackling a robotic task where energy-efficiency may become crucial and controllers can scale, in order to exploit spatiotemporal context and commonsense understanding.

In the scientific domain, this work opens up a fascinating possibility for artificial networks to be constrained by the structure of their biological counterparts.  Alongside research on the computational mechanisms of biological learning \cite{Lansdell253351}, such efforts can introduce new push-pull dynamics between robotics and neuroscience by exploring and exploiting \textit{interpretable} connections between neurophysiology and behavior. Despite being equipped with biologically realistic models of neurons, current SNN can only offer weak suggestions on the underlying neural mechanisms that give rise to the targeted behavior. Recent applications of gradient-descent alternatives to SNNs \cite{RN255, RN28} promise to introduce SNNs to scalable problems, but they inherit the main limitations that deep networks have. For example, any backprop-type learning seeks to match the network’s input to its output, much regarding and, thereby, structuring the network as a black-box. Our bottom-up approach is biologically interpretable and can spur the development of neural-controlled robots as test-beds for understanding how brain function relates to neural structure.  

\section{Conclusion}
Overall, the paper introduces an alternative approach towards designing robot controllers, that of developing SNNs inspired by the brain topology associated with the targeted behavior. We showed how target tracking can be achieved by emulating at a reasonable scale the connectome and the underlying types of neurons, which eliminated the need for training the network. This suggests that building neuro-inspired controllers for this and other types of autonomous robotic behavior is a direction worth pursuing.

\bibliographystyle{IEEEtran}
\bibliography{ref}
\end{document}